\documentclass[12 pt]{article}
\usepackage{amsthm}
\usepackage{amssymb}
\usepackage{amsmath}
\usepackage{geometry}
\usepackage{graphicx}
\usepackage{xcolor}
\usepackage{todonotes, subfigure}
\usepackage{mathrsfs}   
\usepackage[utf8]{inputenc}
\theoremstyle{plain} 
\newtheorem{thm}{Teorema}[section]

\newtheorem{prop}[thm]{Proposizione} 

\theoremstyle{definition} 
\newtheorem{defn}{Definition} 

\theoremstyle{remark}

\newcommand{\bs}{\boldsymbol}

\begin{document}
	
	\title{\bf Persistence kernels for classification \\ A comparative study}
	\author{Cinzia Bandiziol\thanks{cinzia.bandiziol@phd.unipd.it},$\;$ Stefano De Marchi\thanks{stefano.demarchi@unipd.it}}
	\date{Dipartimento di Matematica "Tullio Levi-Civita" \\ University of Padova}
	
	\maketitle
	
	
	\textbf{Abstract}: The aim of the present work is a comparative study of different persistence kernels applied to various classification problems. After some necessary preliminaries on homology and persistence diagrams, we introduce five different kernels that are then used to compare their performances of classification on various datasets. We also provide the Python codes for the reproducibility of results.
	
	\section{Introduction}
	In the last two decades, with the increasing need to analyze big amounts of data, which are usually complex and of high dimension, it was revealed meaningful and helpful to discover further methodologies to provide new information from data. This has brought to the birth of Topological Data Analysis (TDA), whose aim is to extract intrinsic, topological features, related to the so-called "shape of data". Thanks to its main tool, Persistent Homology (PH), it can provide new qualitative information that it would be impossible to extract in any other way. These kinds of features that can be collected in the so-called Persistence Diagram (PD), have been winning in many different applications, mainly related to applied science, improving the performances of models or classifiers, as in our context. Thanks to the strong basis of algebraic topology behind it, the TDA is very versatile and can be applied to data with a priori any kind of structure, as we will explain in the following. This is the reason why there is a wide range of fields of applications like chemistry \cite{chemistry}, medicine \cite{medicine}, neuroscience \cite{Pachauri}, \cite{Brain}, finance \cite{finance} and computer graphics \cite{surface} only to name a few.
	
	An interesting and relevant property of this tool is its stability to noise \cite{Cohen}, which is a meaningful aspect for applications to real-world data. On the other hand, since the space of PDs is only metric one, to use methods that require data to live in a Hilbert space, such as SVM and PCA, it is necessary to introduce the notion of kernel or better Persistence Kernel (PK) that maps PD to space with more structure where it is possible to apply techniques that need a proper definition of inner product.
	
	The goals of the present paper are: first we investigate how to choose values for parameters related to different kernels, then we collect tools for computing PD starting from different kinds of data and finally we compare performances of the main kernels in the classification context. As far as we know, the content of this study is not already present in literature.
	
	The paper is organized as follows: in Section 2 we recall the basic notion related to persistent homology, in Section 3 we describe the problem of classification and solve it using Support Vector Machine (SVM), Section 4 lists the main PK available in literature, Section 5 collects all numerical tests that we have run and in Section 6 we outline conclusions.

	\section{Persistent Homology}
	This brief introduction does not claim to be exhaustive therefore we invite interested readers to refer, for instance, to the works \cite{Fomenko}, \cite{IntroAlgTop}, \cite{Edel1}, \cite{Edel2} and \cite{Cech}. The first ingredient needed is the concept of {\it filtration.} The most common choice in applications is to consider a function $f: \textit{X} \to \mathbb{R}$, where $\textit{X}$ is a topological space that varies based on different contexts, and then take into account the filtration based on the sub-level set given by $f^{-1}(-\infty, a)$, $a \in \mathbb{R}$. For example, such an $f$ can be chosen as the distance function in the case of point cloud data, the gray-scale values at each pixel for images, the heat kernel signature for datasets as SHREC14 \cite{SHREC14}, the weight function of edges for graphs, and so on. We now recall the main theoretical results related to point cloud data but all of them can be easily apply to other contexts.
	
	We assume to have a set of points $\mathcal{X} = \{\textbf{x}_{k}\}_{k=1,...,m}$ that, we suppose to live in an open set of a manifold $\mathcal{M}$. The aim is to be able to capture relevant intrinsic properties of the manifold itself and this is achieved through Persistent Homology (PH) applied to such discrete information. To understand how PH has been introduced, first, we have to mention the simplicial homology, which represents the extension of homology theory to structures called simplicial complexes.
	
	\begin{figure}[h]
		\centering
		\includegraphics[scale=0.95]{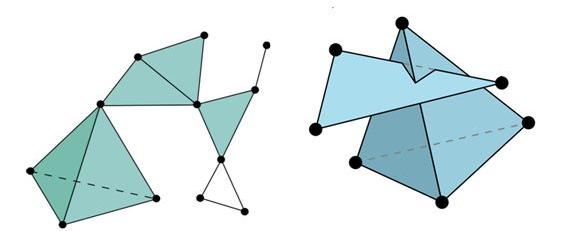}
		
		\caption{An example of a valid simplicial complex (left) and an invalid one (right)}
	\end{figure}
	
	\begin{defn}
		A \textbf{simplicial complex} $K$ consists of a set of simplices of different dimensions and has to meet the following conditions:
		\begin{itemize}
			\item Every face of a simplex $\sigma$ in $K$ must belong to $K$
			\item The non-empty intersection of any two simplices $\sigma _{1},\sigma _{2} \in K$ is a face of both $\sigma _{1}$ and $\sigma _{2}$
		\end{itemize}
		The dimension of $K$ is the maximum dimension of simplices that belong to $K$.
	\end{defn}
	
	In application, data analysts usually compute the {\it Vietoris-Rips complex}.
	
	\begin{defn}
		Let $(\mathcal{X},d)$ denote a metric space from which the samples are taken. The \textbf{Vietoris-Rips complex} for $\mathcal{X}$, associated to the parameter $\epsilon$, denoted by $VR(\mathcal{X}, \epsilon)$, is the simplicial complex whose vertex set is $\mathcal{X}$ and $\{\textbf{x}_{0},\dots,\textbf{x}_{k}\}$ spans a k-simplex if and only if $d(\textbf{x}_{i},\textbf{x}_{j}) \leqslant 2 \epsilon$ for all $0 \leqslant i,j \leqslant k$.
	\end{defn}
	
	If $K:=VR(\mathcal{X}, \bar{\epsilon})$, we can divide all simplices of this set $K$, into groups based on their dimension $k$ and we can enumerate them using $\Delta_{i}^{k}$. If $G=(\mathbb{Z},+)$ is the well known Abelian group, we may build linear combinations of simplices with coefficients in G, and so we introduce the following
	
	\begin{defn}
		An object of the form $c = \sum_{i}a_{i}\Delta^{k}_{i}$ with $a_{i} \in \mathbb{Z}$ is a \textbf{integer valued k-dimensional chain}.
	\end{defn}
	
	Linearity allows to extend the previous definition to any subsets of simplices of $K$ with dimension $k$,
	
	\begin{defn}
		The group $C_{k}^{\bar{\epsilon}}(\mathcal{X})$ is called the \textbf{group of k-dimensional simplicial integer-valued chains} of the simplicial complex $K$.
	\end{defn}
	
	It is then possible to associate to each simplicial complex, the corresponding set of Abelian groups $C_{0}^{\bar{\epsilon}}(\mathcal{X}),\dots,C_{n}^{\bar{\epsilon}}(\mathcal{X})$.
	
	\begin{defn}
		The boundary $\partial \Delta^{k}$ of an oriented simplex $\Delta^{k}$, is the sum of all its $(k-1)$-dimensional faces taken with a chosen orientation. More precisely
		
		$$\partial \Delta^{k} = \sum_{i=0}^{k} (-1)^{k} \Delta^{k-1}_{i}.$$
		
	\end{defn}

	In the general setting, we can extend the boundary operator by linearity to a general element of $C_{k}^{\bar{\epsilon}}(\mathcal{X})$, obtaining a map $\partial_{k}: C_{k}^{\bar{\epsilon}}(\mathcal{X}) \to C_{k-1}^{\bar{\epsilon}}(\mathcal{X})$.
	
	No matter what the value of $k$ is, it is a linear map. Therefore we can take into account its kernel, for instance, the group of $k$-cycles, $Z_{k}^{\bar{\epsilon}}(\mathcal{X}):=\ker(\partial_{k})$ and the image, the group of $k$-boundaries, $B_{k+1}^{\bar{\epsilon}}(\mathcal{X}):=\mathrm{im}(\partial_{k})$. Then $H_{k}^{\bar{\epsilon}}(\mathcal{X}) = Z_{k}^{\bar{\epsilon}}(\mathcal{X})/B_{k+1}^{\bar{\epsilon}}(\mathcal{X})$ is the k-homology group and represents the k-dimensional holes that can be recovered from the simplicial structure.
	We briefly recall here that, for instance, 0-dimensional holes correspond to connected components, 1-dimensional holes are cycles, and 2-dimensional holes are cavities/voids. Since they are algebraic invariants, they collect qualitative information regarding the topology of the data. The most crucial aspect is highlighting the best value for $\epsilon$ to obtain a simplicial complex $K$ that faithfully reproduces the original manifold's topological structure. The answer is not straightforward and the process reveals unstable, therefore the PH analyzes not only one simplicial complex but a nested sequence of them, and, following the evolution of such structure, it notes down features that gradually emerge. From a theoretical point of view, letting $0 < \epsilon_{1} < \dots < \epsilon_{l}$ be an increasing sequence of real numbers, we obtain the \textbf{filtration}
	
	$$\emptyset \subset K_{1} \subset K_{2} \subset \dots \subset K_{l}$$
	
	with $K_{i} = VR(\mathcal{X},\epsilon_{i})$ and then
	
	\begin{defn}
		The \textbf{p-persistent homology group} of $K_{i}$ is the group defined as
		$$H_{k}^{i,p} = Z_{k}^{i} / (B_{k}^{i+p} \cap Z_{k}^{i})$$
		
	\end{defn}
	
	This group contains all stable homology classes in the interval $i$ to $i+p$: they are born before the time/index i and are still alive after p steps. The persistent homology classes alive for large values of $p$ are stable topological features of $\mathcal{S}$ (see \cite{TDA_FMNIST}). Along the filtration, the topological information appears and disappears, thus it means that they may be represented with a couple of indexes. If p is such a feature, it must be born in some $K_{i}$ and die in $K_{j}$ so it can be described as $(i,j)$, $i < j$. We underline here that $j$ can be equal to $+\infty$, since some features can be alive up to the end of the filtration. Hence, all such topological invariants live in the extended positive plane, that here is denoted by $\mathbb{R}^{2}_{+}=\mathbb{R}_{\geq 0} \times \{\mathbb{R}_{\geq 0} \cup \{+ \infty\}\}$. Another interesting aspect to highlight is that some features can appear more than once and accordingly such collection of points are called multisets. All of these observations are grouped into the following
	
	\begin{defn}
		A \textbf{Persistence Diagram (PD)} \textit{D}$_{r}(\mathcal{X}, \bs{\varepsilon})$ related to the filtration $\emptyset \subset K_{1} \subset K_{2} \subset \dots \subset K_{l}$ with $\bs{\varepsilon} := (\epsilon_{1},\dots,\epsilon_{l})$ is a multiset of points defined as
		
		$$\textit{D}_{r}(\mathcal{X},\bs{\varepsilon}) := \{(b,d)| (b,d) \in P_{r}(\mathcal{X},\bs{\varepsilon} )\} \cup \Delta$$
		
		where $P_{r}(\mathcal{X},\bs{\varepsilon} )$ denotes the set of $r$-dimensional birth-death couples that came out along the filtration, each $(b,d)$ is considered with its multiplicity, while points of $\Delta = \{(x,x)|x \geq 0\}$ with infinite multiplicity. One may consider all $P_{r}(\mathcal{X},\bs{\varepsilon})$ for every $r$ together, obtaining the total PD denoted here by $\textit{D}(\mathcal{X},\bs{\varepsilon})$, that we will usually consider in the following sections.
		
		Each point $(b,d) \in \textit{D}_{r}(\mathcal{X},\bs{\varepsilon})$ is called \textbf{generator} of the persistent homology, and represents a topological property which appears at $K_{b}$ and disappears at $K_{d}$. The difference $d-b$ is called \textbf{persistence} of the generator, represents its lifespan and shows the robustness of the topological property.
	\end{defn}
	
	\begin{figure}[h]
		\centering
		\includegraphics[scale=0.75]{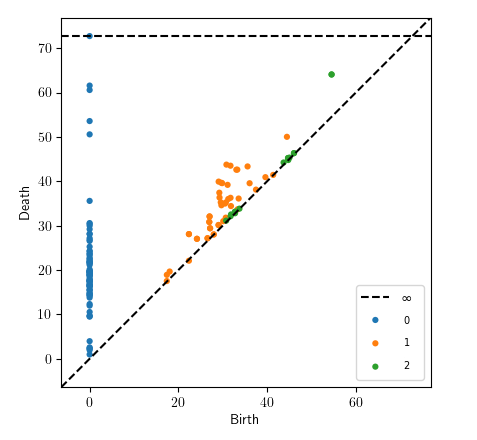}
		\caption{An example of Persistence Diagram with features of dimensions 0,1 and 2}
		\label{PD}
	\end{figure}
	
	Figure \ref{PD} is an example of total PD collecting features of dimension 0 (in blue), of dimension 1 (in orange), and of dimension 2 (in green). Points close to the diagonal represent features with a short lifetime, and so usually they are concerned with noise, instead features far away are indeed relevant and meaningful and, based on applications, one can decide to consider both or only the most interesting ones. At the top of the Figure, there is a dashed line that indicates the infinity and allows to plot also couples as $(i, +\infty)$.
	
	In the previous definition, the set $\Delta$ is added to finding out proper bijections between sets, that without $\Delta$ could not have the same number of points. It makes it possible to compute the proper distance between PDs. 
	
	\subsection{Stability}
	A key property of PDs is stability under perturbation of the data. First, we recall two famous distances for sets,
	
	\begin{defn}
		For two nonempty sets $\mathcal{X},\mathcal{Y} \subset \mathbb{R}^{d}$ with the same cardinality, the \textbf{Haussdorff distance} is  
		
		$$d_{H}(\mathcal{X},\mathcal{Y}) := \max \{\sup_{x \in \mathcal{X}} \inf_{y \in \mathcal{Y}} \|x-y\|_{\infty}, \sup_{y \in \mathcal{Y}} \inf_{x \in \mathcal{X}} \|y-x\|_{\infty}\}$$
		
		and \textbf{bottleneck distance} is defined as
		
		\begin{equation}
			d_{B}(\mathcal{X},\mathcal{Y}) := \inf_{\gamma}\sup_{x \in \mathcal{X}} \|x-\gamma(x)\|_{\infty} \label{Bott}
		\end{equation}
		
		where we consider all possible bijection of multisets $\gamma:\mathcal{X} \rightarrow \mathcal{Y}$. Here, we use
		
		$$\|v-w\|_{\infty} = \max\{|v_{1}-w_{1}|,|v_{2}-w_{2}|\},\ \ \text{for } v = (v_{1},v_{2}), w = (w_{1},w_{2})\in \mathbb{R}^{2}$$
	\end{defn}
	
	\begin{figure}[h]
		\centering
		\includegraphics[scale=0.65]{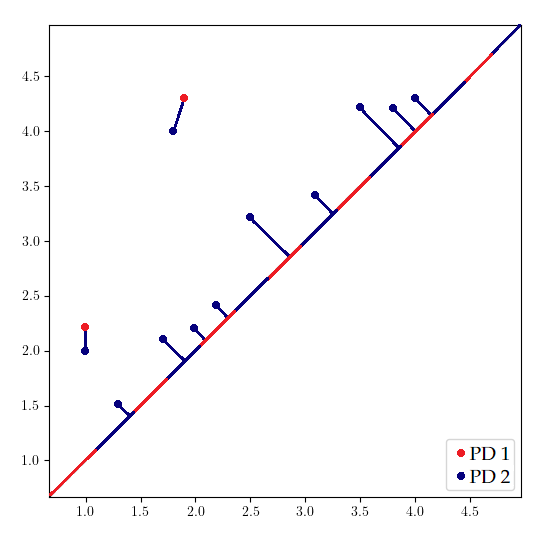}	
		\caption{Example of bottleneck distance between two PDs in red and blue}\label{bott_fig}
	\end{figure}
	
	We try to explain better how to compute the bottleneck distance. We have to take all possible ways to move points from $\mathcal{X}$ to $\mathcal{Y}$ in a bijective manner and then one can compute properly the distance. Figure \ref{bott_fig} shows two different PDs overlapped, that consist of $\Delta$ joined with 2 points in red and 11 points in blu respectively. First, in order to apply the definition (\ref{Bott}), we need two sets with same cardinality. For this aim, it is necessary to add points of $\Delta$, more precisely the orthogonal projection onto the diagonal of the  9 blue points closer to it, to reach 11. Lines between points and $\Delta$ represent the bijection that realizes the best matching between points in definition (\ref{Bott}).
	
	\begin{prop}
		Let $\mathcal{X}$ and $\mathcal{Y}$ be finite subset in a metric space $(M,d_{M})$. Then the Persistence Diagrams $D(\mathcal{X}, \bs{\varepsilon})$, $D(\mathcal{Y}, \bs{\varepsilon})$ satisfy
		
		$$d_{B}(D(\mathcal{X}, \bs{\varepsilon}),D(\mathcal{Y},\bs{\varepsilon})) \leqslant d_{H}(\mathcal{X},\mathcal{Y}).$$
	\end{prop}
	
	For any further details see for example \cite{IntroAlgTop}.
	
	\section{SVM}
	\subsection{Classification problem}
	Let $\Omega \subset \mathbb{R}^{d}$ and $ \{ \textbf{x}_{1},...,\textbf{x}_{m}\} \subset \mathcal{X} \subset \Omega$ be the set of input data with $d,m \in \mathbb{N}.$ We have a training set, composed by the couples $(\textbf{x}_{i},y_{i})$ with $i=1,...,m$ and $y_{i} \in \mathcal{Y} = \{-1,1\}$. The \textbf{binary supervised learning task} consists in finding a function $f: \Omega \longrightarrow \mathcal{Y}$, the model, such that it can predict, in a satisfactory way, the label of an unseen $\tilde{\textbf{x}} \in \Omega \setminus \mathcal{X}$.
	
	The aim is to find the hyperplane that can separate, in the best possible way, points that belong to different classes and from here the name separating hyperplane. The best possible way means that it separates the two classes with the higher margin, that is the distance between the hyperplane and the points of both classes.
	
	More formally, if we assume to be in a space $\mathscr{F}$ with dot product, for instance $\mathscr{F}$ can be a subset of $\mathbb{R}^{d}$ with $\langle \cdot,\cdot\rangle$, since a generic hyperplane can be defined as
	
	$$\{\textbf{x} \in \mathscr{F} | \langle \textbf{w},\textbf{x}\rangle + b = 0\} \text{ } w \in \mathscr{F}, b \in \mathbb{R}$$
	
	one can introduce
	
	\begin{defn}
		We call
		
		$$\rho_{w,b}(\textbf{x},y) := \frac{y(\langle w,\textbf{x} \rangle  + b)}{\|w\|}$$
		
		the \textbf{geometrical margin} of the point $(\textbf{x},y) \in \mathscr{F} \times \{-1,1\}$. Instead, the minimum value,
		
		$$\rho_{w,b} := \min_{i=1,\dots,m} \rho_{w,b}(\textbf{x}_{i},y_{i})$$
		
		shall be call the \textbf{geometrical margin} of $(\textbf{x}_{1},y_{1}),\dots,(\textbf{x}_{m},y_{m})$.
	\end{defn}
	
	From a geometrical point of view, this margin measures effectively the distance between samples and the hyperplane itself. Then SVM is looking for a suitable hyperplane that, intuitively realizes the maximum of such margin. For any further details, see for example \cite{LWK}.
	The precise formalization brings to an optimization problem that, thanks to the Lagrange multipliers and Karush-Kuhn-Tucker conditions, it turns out to have the following formulation, as \textbf{SVM optimization problem},
	\begin{align*}
		\max_{\alpha\ \in\ \mathbb{R}^{m}} \ \ \ &\sum_{i=1}^{m} 	\alpha_{i} - \frac{1}{2}\sum_{i,j=1}^{m} \alpha_{i}\alpha_{j} y_{i}y_{j}\langle \textbf{x}_{i},\textbf{x}_{j}\rangle\\
		\text{s. to\ \ \ \ } &\sum_{i=1}^{m} \alpha_{i}y_{i}=0\\
		& 0 \leq \alpha_{i} \leq C \ \forall i = 1,\dots,m
	\end{align*}
	where $[0,C]^{m}$ is the bounding box $C \in [0,+\infty)$ and $\alpha_{i}>0$ are called \textbf{Support Vectors}. From here and what follows the name Support Vector Machine is shortened in SVM and $\langle \cdot, \cdot \rangle$ denotes the inner product in $\mathbb{R}^{d}$.
	This formulation can face satisfactorily the classification task if data are linearly separable. In applications, it doesn't happen frequently and so it is needed to introduce some nonlinearity and move in higher dimensional space where, hopefully, that can happen. This can be achieved with the use of kernels.
	Starting from the original dataset $\mathcal{X}$, the theory tells to introduce a feature map $\Phi: \mathcal{X} \to \mathcal{H}$ that moves data from $\mathcal{X}$ to a Hilbert space of function $\mathcal{H}$, the so-called \textbf{feature space}. The kernel is then defined as $\kappa(\textbf{x},\bar{\textbf{x}}):=\langle \Phi(\textbf{x}),\Phi(\bar{\textbf{x}}) \rangle_{\mathcal{H}}$ (\textbf{kernel trick}). Thus the optimization problem becomes
	
	\begin{align*}
		\max_{\alpha\ \in\ \mathbb{R}^{m}} \ \ \ &\sum_{i=1}^{m} \alpha_{i} - \frac{1}{2}\sum_{i,j=1}^{m} \alpha_{i}\alpha_{j} y_{i}y_{j}\kappa(\textbf{x}_{i},\textbf{x}_{j})\\
		\text{s. to\ \ \ \ } &\sum_{i=1}^{m} \alpha_{i}y_{i}=0\\
		& 0 \leq \alpha_{i} \leq C \ \forall i = 1,\dots,m
	\end{align*}
	
	where kernel represents a generalization of the inner product in $\mathbb{R}^{d}$. We are interested in classifying PDs and obviously, we need suitable definitions for kernels for PDs, the so-called \textbf{Persistence Kernels (PK)}.
	
	\section{Persistence Kernels}
	
	In what follows we denote with $\mathfrak{D}$ the set of the total PDs.
	
	\subsubsection{Persistence Scale-Space Kernel (PSSK)}
	
	The first kernel was described in \cite{PSSK}. The main idea is to compute the feature map as the solution of the Heat equation. We consider $\Omega_{ad} = \{\textbf{x} = (x_{1},x_{2}) \in \mathbb{R}^{2} : x_{2} \geqslant x_{1}\}$ and we denote with $\delta_{\textbf{x}}$ the Dirac delta centered at \textbf{x}. For a given $D \in \mathfrak{D}$, we consider the solution $u: \Omega_{ad} \times \mathbb{R}_{\geqslant 0} \rightarrow \mathbb{R}$, $(\textbf{x},t) \mapsto u(\textbf{x},t)$ of the following PDE:	
	\begin{align*}
		\Delta_{\textbf{x}}u=\partial_{t}u \ \ \ & \text{in} \ \Omega_{ad} \times \mathbb{R}_{\geqslant 0}\\
		u = 0 \ \ \ & \text{on} \ \partial\Omega_{ad} \times \mathbb{R}_{\geqslant 0}\\
		u = \sum_{\textbf{y} \in D} \delta_{\textbf{y}} \ \ \ & \text{on} \  \Omega_{ad} \times 0.
	\end{align*}
	
	The feature map $\Phi_{\sigma}: \mathfrak{D} \rightarrow \mathrm{L}^{2}(\Omega_{ad})$ at scale $\sigma > 0$ at $D$ is defined as $\Phi_{\sigma}(D) = u\big|{}_{t=\sigma}$. This map yields the \textbf{Persistence Scale-Space Kernel} (\textbf{PSSK}) $K_{PSS}$ on $\mathfrak{D}$ as:
	$$K_{PSS}(D,E)=\langle \Phi_{\sigma}(D),\Phi_{\sigma}(E)\rangle_{\mathrm{L}^{2}(\Omega_{ad})}.$$
	
	But since it is known as an explicit formula for the solution $u$, the kernel takes the form
	$$K_{PSS}(D,E) = \frac{1}{8 \pi \sigma} \sum_{\textbf{x} \in D, \textbf{y} \in E} \exp (-\frac{\|\textbf{x}-\textbf{y}\|^{2}}{8\sigma})-\exp (-\frac{\|\textbf{x}-\bar{\textbf{y}}\|^{2}}{8\sigma})$$
	where $\textbf{y}=(a,b) \text{, } \bar{\textbf{y}}=(b,a)$, for any $D,E \in \mathfrak{D}$.
	
	\subsubsection{Persistence Weighted Gaussian Kernel (PWGK)}
	
	In \cite{PWGK}, the authors introduce a new kernel whose idea is to replace each PD with a discrete measure. Starting with a strictly positive definite kernel, as for example the gaussian one $\kappa_{G}(\textbf{x},\textbf{y}) = e^{-\frac{\|\textbf{x}-\textbf{y}\|^{2}}{2 \sigma^{2}}}$, $\sigma > 0$ we denote the corresponding Reproducing Kernel Hilbert Space $\mathcal{H}_{\kappa_{G}}$.
	
	If $\Omega \subset \mathbb{R}^{d}$, we denote with $M_{b}(\Omega)$ the space of finite signed Radon measures and
	$$E_{\kappa_{G}}: M_{b}(\Omega) \to \mathcal{H}_{\kappa_{G}}, \mu \mapsto \int_{\Omega} \kappa_{G}(\cdot, \textbf{x}) d\mu(\textbf{x}).$$
	
	For any $D \in \mathfrak{D}$, if $\mu_{D}^{w} = \sum_{\textbf{x} \in D} w(\textbf{x}) \delta_{\textbf{x}}$, where the weight function satisfies $w(x) > 0$ for all $\textbf{x} \in D$ then
	$$E_{\kappa_{G}}(\mu_{D}^{w}) =\sum_{\textbf{x} \in D} w(\textbf{x})\kappa_{G}(\cdot,\textbf{x})$$
	
	where
	
	$$w(\textbf{x})=\arctan(C_{w}pers(\textbf{x})^{p})$$
	
	and $pers(\textbf{x}) = x_{2}-x_{1}$.
	
	The \textbf{Persistence Weight Gaussian Kernel} (\textbf{PWGK}) is defined as
	$$K_{PWG}(D,E) = \exp\bigg(-\frac{1}{2 \tau^{2}} \|E_{\kappa_{G}}(\mu_{D}^{w})-E_{\kappa_{G}}(\mu_{E}^{w}) \|^{2}_{\mathcal{H}_{\kappa_{G}}}\bigg) \text{, } \tau > 0$$
	for any $D,E \in \mathfrak{D}$.
	
	\subsubsection{Sliced Wasserstein Kernel (SWK)}
	Another possible choice for $\kappa$ has been introduced in \cite{SWK}.
	
	We consider $\mu$ and $\nu$ two nonnegative measures on $\mathbb{R}$ such that $\mu(\mathbb{R}) = r = |\mu|$ and $\nu(\mathbb{R}) = r = |\nu|$, we recall that the 1-Wasserstein distance for nonnegative measures is defined as
	$$\mathcal{W}(\mu,\nu) = \inf_{P \in \Pi(\mu,\nu)} \int \int_{\mathbb{R} \times \mathbb{R}} |x-y| dP(x,y)$$
	
	where $\Pi(\mu,\nu)$ is the set of measures on $\mathbb{R}^{2}$ with marginals $\mu$ and $\nu$.
	
	\begin{defn}
		Given $\theta \in \mathbb{R}^{2}$ with $\|\theta\|_{2}=1$, let $L(\theta)$ denote the line $\{\lambda \theta | \lambda \in \mathbb{R}\}$ and let $\pi_{\theta}:\mathbb{R}^{2} \to L(\theta)$ be the orthogonal projection onto $L(\theta)$. Let $D,E \in \mathfrak{D}$ and let $\mu_{D}^{\theta}:=\sum_{\textbf{x} \in D} \delta_{\pi_{\theta}(\textbf{x})}$ and $\mu_{D\Delta}^{\theta}:=\sum_{\textbf{x} \in D} \delta_{\pi_{\theta}\circ\pi_{\Delta}(\textbf{x})}$ and similarly for $\mu_{E}^{\theta}$ and $\mu_{E\Delta}^{\theta}$ where $\pi_{\Delta}$ is the orthogonal projection onto the diagonal. Then, the \textbf{Sliced Wasserstein distance} is
		$$SW(D,E) = \frac{1}{2 \pi} \int_{\mathbb{S}_{1}} \mathcal{W}(\mu_{D}^{\theta}+\mu_{E\Delta}^{\theta},\mu_{E}^{\theta}+\mu_{D\Delta}^{\theta}) d\theta.$$
	\end{defn}
	
	Thus, the \textbf{Sliced Wasserstein Kernel} (\textbf{SWK}) is defined as
	
	$$K_{SW}(D,E) := \exp\bigg(-\frac{SW(D,E)}{2\eta^{2}}\bigg) \text{, $\eta > 0$}$$
	
	for any $D,E \in \mathfrak{D}$.

	\subsubsection{Persistence Fisher Kernel (PFK)}
	In \cite{Fisher}, the authors describe a kernel based on Fisher Information geometry.
	
	A persistence diagram $D \in \mathfrak{D}$ can be considered as a discrete measure $\mu_{D}=\sum_{u \in D} \delta_{u}$, where $\delta_{u}$ is the Dirac's delta centered in u. Given a bandwidth $\sigma > 0$, and a set $\Theta$, one can smooth and normalize $\mu_{D}$ as follows
	
	$$\rho_{D} := \frac{1}{Z} \sum_{u \in D} N(\textbf{x};u,\sigma I)$$ 
	
	where $N$ is a Gaussian function, $Z = \int_{\theta}\sum_{u \in D} N(x;u,\sigma I) dx$ and $I$ is the identity matrix. Thus, using this measure, any PD can be regarded as a point in $\mathbb{P} = \{\rho | \int\rho(\textbf{x})d\textbf{x} = 1, \rho(\textbf{x}) \geq 0\}$.
	
	Given two element in $\rho_{i},\rho_{j} \in \mathbb{P}$, the \textbf{Fisher Information Metric} is
	
	$$d_{\mathbb{P}}(\rho_{i},\rho_{j}) = \arccos\bigg(\int \sqrt{\rho_{i}(\textbf{x})\rho_{j}(\textbf{x})} d\textbf{x}\bigg).$$
	
	Inspiring by the Sliced Wasserstein Kernel construction, we have the following
	
	\begin{defn}
		Let $D, E$ be two finite and bounded persistence diagrams. The Fisher information metric between $D$ and $E$, is defined as
		
		$$d_{FIM} (D,E) := d_{\mathbb{P}}(\rho_{D \cup E_{\Delta}},\rho_{E \cup D_{\Delta}})$$
		
		where $D_{\Delta} := \{\Pi_{\Delta}(u) | u \in D\}$, $E_{\Delta} := \{\Pi_{\Delta}(u) | u \in E\}$ and $\Pi_{\Delta}$ is the orthogonal projection on the diagonal $\Delta = \{(a,a)|a \geq 0\}$.
	\end{defn}
	
	The \textbf{Persistence Fisher Kernel} (\textbf{PFK}) is then defined as
	$$K_{PF}(D,E) := \exp(-t d_{FIM}(D,E)) \text{, } t > 0 \text{, for any $D,E \in \mathfrak{D}$.}$$
	
	\subsubsection{Persistence Image (PI)}
	The main reference is \cite{PI}. If $D \in \mathfrak{D}$ we introduce a change of coordinates, $T:\mathbb{R}^{2} \to \mathbb{R}^{2}$ given by $T(x,y) = (x,y-x)$ and let $T(D)$ be the transformed multiset in first-persistence coordinates. Let $\phi_{u}:\mathbb{R}^{2} \to \mathbb{R}$ be a differentiable probability distribution with mean $u=(u_{x},u_y) \in \mathbb{R}^{2}$, usually $\phi_{u}=g_{u}$, where $g_{u}$ is the 2-dimensional Gaussian with mean $u$ and variance $\sigma^{2}$, defined as
	
	$$g_{u}(x,y) = \frac{1}{2\pi\sigma^{2}}e^{-[(x-u_{x})^{2}+(y-u_{y})^{2}]/2\sigma^{2}}.$$
	
	Fix a weight function $f:\mathbb{R}^{2} \to \mathbb{R}$, that is $f \geq 0$, it is equal zero on the horizontal axis, continuous and piecewise differentiable. A possible choice is a function that depends only to the persistence coordinate y, a function $f(x,y) = w_{b}(y)$ where
	
	\begin{equation*}
		w_{b}(t) = \left\{
		\begin{array}{@{}rl}
			0 & \text{if } t \leq 0,\\
			\frac{t}{b} & \text{if } 0 < t < b,\\
			1 & \text{if } t \geq b.
		\end{array} \right.
	\end{equation*}

	\begin{defn}
		Given $D \in \mathfrak{D}$, the corresponding \textbf{persistence surface} $\rho_{D}:\mathbb{R}^{2} \to \mathbb{R}$ is the function
		
		$$\rho_{D}(x,y) = \sum_{u \in T(D)} f(u)\phi_{u}(x,y).$$
		
	\end{defn}
	
	If we divide the plane in a grid with $n^{2}$ pixels $(P_{i,j})_{i,j=1,\dots,n}$, we have the following
	
	\begin{defn}
		Given $D \in \mathfrak{D}$, its \textbf{persistence image} is the collection of pixels
		
		$$PI(\rho_{D})_{i,j} = \int\int_{P_{i,j}} \rho_{D}(x,y)dxdy.$$
	\end{defn}
	
	Thus, through persistence image, each persistence diagram is turned into a vector $PIV \in \mathbb{R}^{n^{2}}$ that is $PIV(D)_{i+n(j-1)}=PI(D)_{i,j}$, then it is possible to introduce the following kernel
	
	$$K_{PI}(D,E) = <PIV(D),PIV(E)>_{\mathbb{R}^{n^{2}}}.$$

	\section{Shape paramenters analysis}
	Each aforementioned kernel has some parameters, that have been chosen through the cross-validation phase. As in the context of RBF as explained in \cite{Fass-Meshfree}, also in Machine Learning framework, it is better to tune the parameters accordingly to the so-called trade-off principle, for instance, such that the condition number of Gram matrix is not so high and on the other hand the accuracy is satisfactorily high. The aim here is to run such analysis to kernels presented in this paper.
	
	The \textbf{PSSK} has only one parameter to tune $\sigma$. Typically the users consider $\sigma \in \{0.001,0.01,0.1,1,10,100,1000\}$. We run the CV phase for different shuffles of a dataset and plot the results in terms of the condition number of the Gram matrix related to the training samples and the accuracy. For our analysis, we consider $\sigma \in \{0.00001,0.0001,0.001,0.01,0.1,1,10,100,500,800,1000\}$ and run tests on some datasets cited in the following. The results are similar in each case so we decided to report ones about SHREC14 dataset.
	
	\begin{figure}[h]
		\centering
		\includegraphics[scale=0.55]{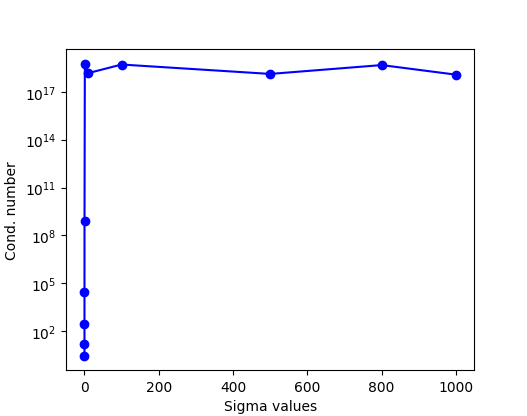} \quad \includegraphics[scale=0.55]{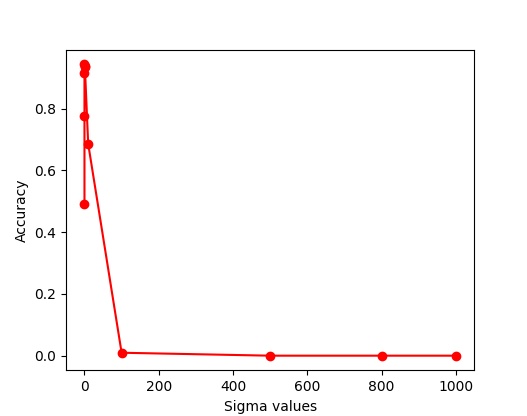}
		
		\caption{Comparison results about PSSK for SHREC14 in terms of condition number (left) and accuracy (right) with different $\sigma$}
	\end{figure}

	From the plot, it is evident how large values of $\sigma$ bring to unstable matrix and less accuracy. Then in what follows, we will take into account only $\sigma \in \{0.00001,0.0001,0.001,$ $0.01,0.1,1,10\}$.
	
	\textbf{PWGK} is the kernel with a higher number of parameters to tune, then it is not so evident what are the best-set values to take into account. We choose reasonable starting sets as: $\tau \in \{0.001,0.01,0.1,1,10,100, 1000\}$, $\rho \in \{0.001, 0.01, 0.1,1,10,100,1000\}$, $p \in \{1,5,10,50,100\}$, $C_{w} \in \{0.001,0.01,0.1,1\}$. Due to a large number of parameters, we first ran some experiments varying $(\rho, \tau)$ with fixed $(p,C_w)$, and then we reversed the roles.
	
	\begin{figure}[h]
		\begin{center}
			\includegraphics[scale=0.35]{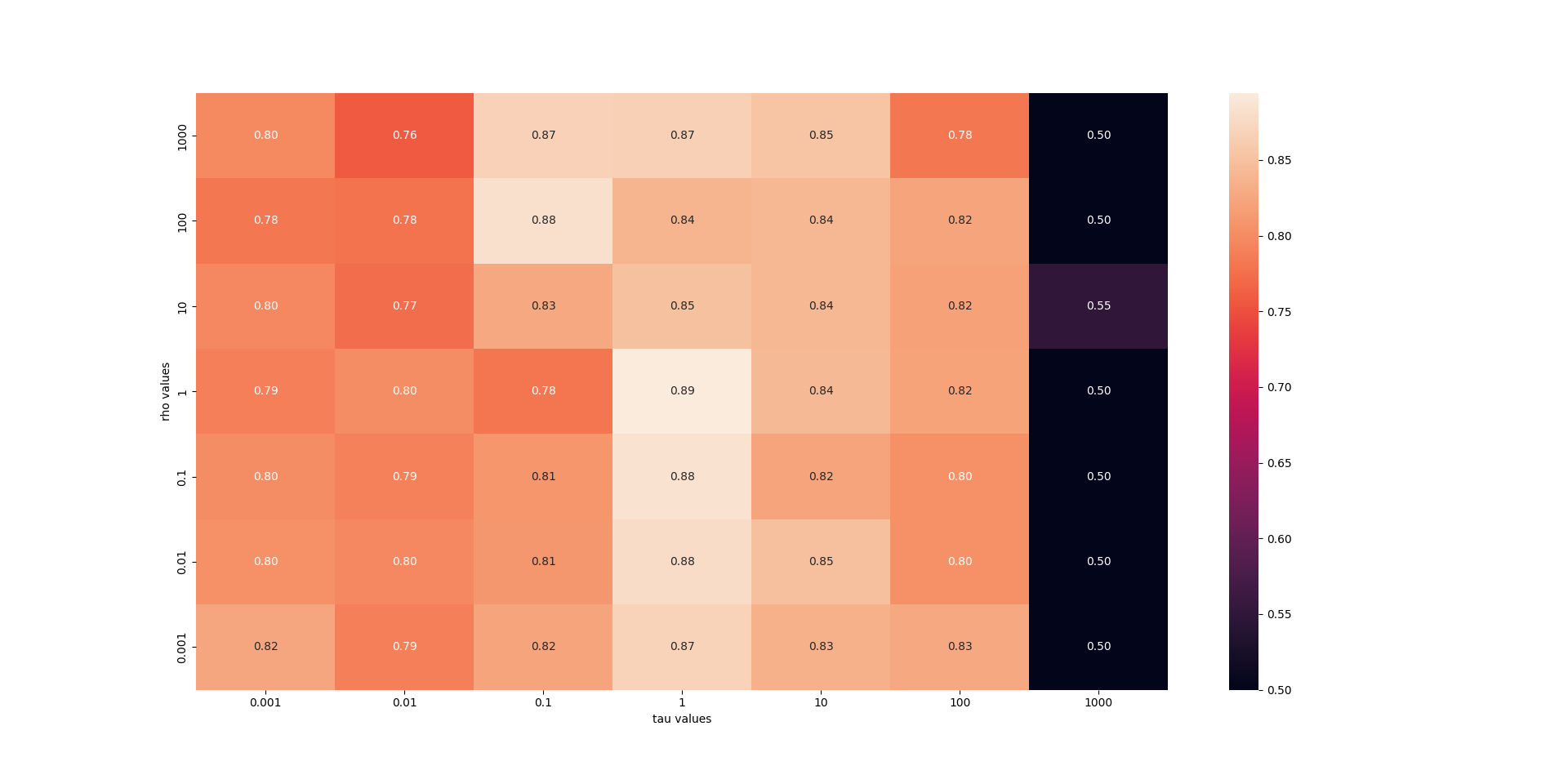}
		\end{center}\caption{Comparison results about PWGK for MUTAG in terms of accuracy with different $\tau$ and $\rho$} \label{PWGK}
	\end{figure}
	
	We report here in Figure \ref{PWGK} only a plot for fixed $C_{w}$ and $p$ because it highlights how high values of $\tau$ (for example $\tau = 1000$) are to be excluded. We find this behavior for different values of $C_{w},p$ and various datasets, here the case $C_{w}=1$, $p=10$ and MUTAG dataset. 
	Therefore we decide to vary the parameters as follows: $\tau \in \{0.001,0.01,0.1,1,10,100\}$, $\rho \in \{0.001, 0.01, 0.1,1,10,100,1000\}$, $p \in \{1,5,10,50,100\}$, $C_{w} \in \{0.001,0.01,0.1,1\}$. Unluckily there is no other evidence that can guide the choices, except for $\tau$, where values $\tau=1000$ always have bad accuracy, as one can see below in the case of MUTAG with shortest path distance.
	
	\newpage
	
	In the case of \textbf{SWK}, there is only one parameter $\eta$. In \cite{SWK}, the authors propose to consider values starting from the first and last decile and the median value of the gram matrix of the training samples flatten in order to obtain a vector, then they multiply these three values for $0.01,0.1,1,10,100$. For our analysis, we have decided to study the behavior of such kernel considering the same set of values, independently from the specific dataset. We consider $\eta \in \{0.00001,0.0001,0.001,0.01,0.1,1,10,100,500,800,1000\}$.
	
	\begin{figure}[h]
		\centering
		\includegraphics[scale=0.45]{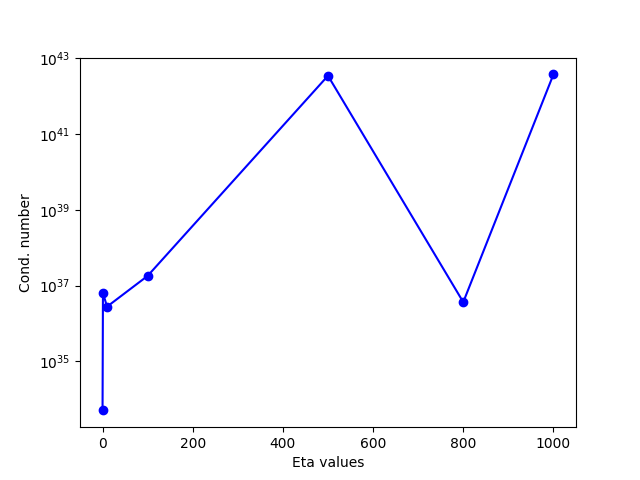} \quad \includegraphics[scale=0.45]{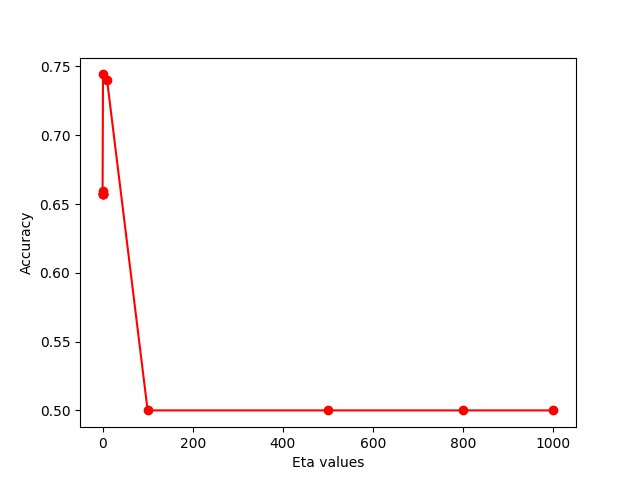}
		
		\caption{Comparison results about SWK for DHFR in terms of condition number (left) and accuracy (right) with different $\eta$}
	\end{figure}
	
	We run tests on some datasets and the plot, related to the DHFR dataset, reveals evidently that large values for $\eta$ are to be excluded. So, we will decide to take $\eta$ only in $\{0.00001,0.0001,0.001,0.01,0.1,1,10\}$.
	
	\begin{figure}[h]
		\begin{center}
			\includegraphics[scale=0.45]{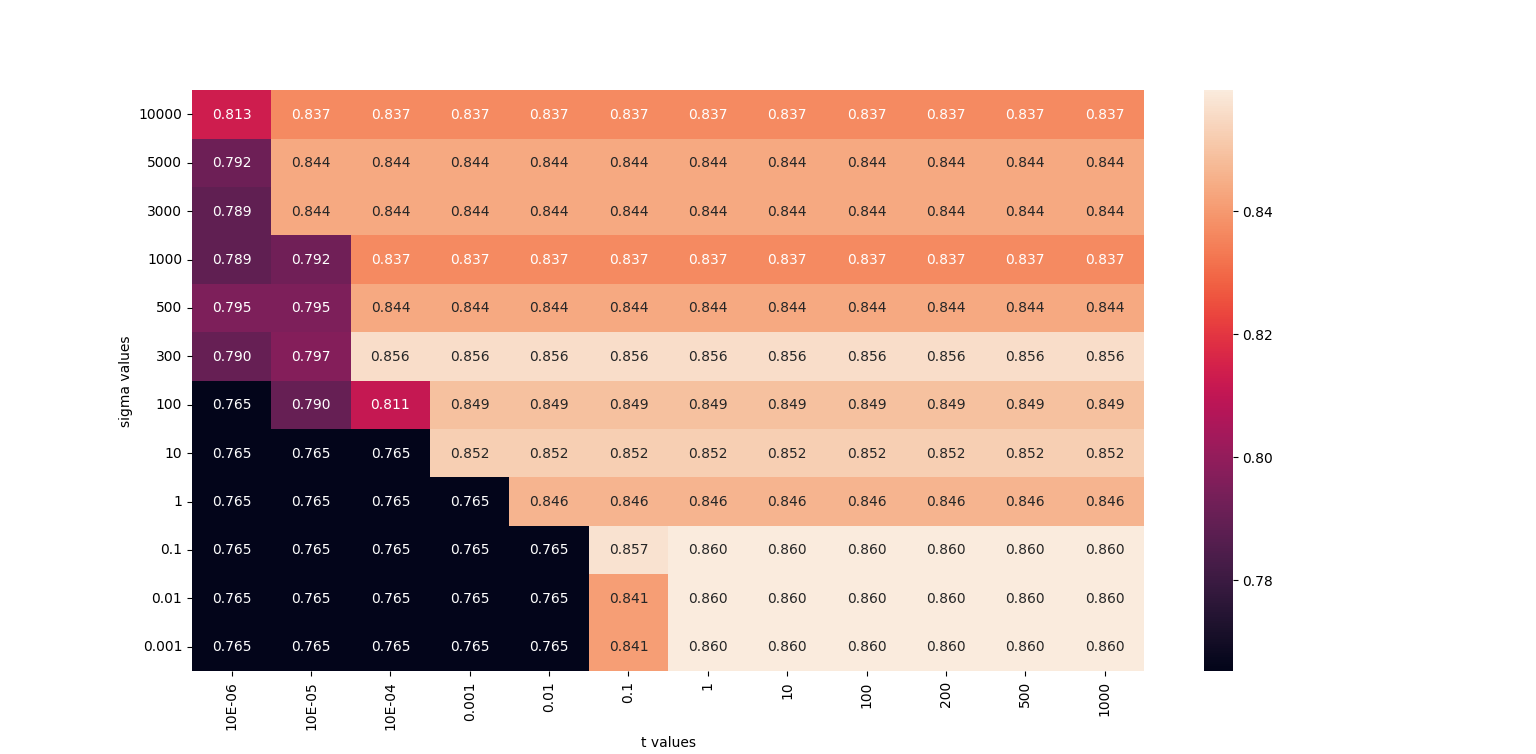}
		\end{center}\caption{Comparison results about PFK for MUTAG in terms of accuracy with different t and $\sigma$}
	\end{figure}		
	
	\textbf{PFK} has two parameters: the variance $\sigma$ and $t$. In \cite{Fisher}, the authors exhibit the procedure to follow in order to obtain the corresponding set of values. It shows that the choice of t depends on $\sigma$. Instead, our aim in this paper is to carry out an analysis that is dataset-independent and that turns out to be strictly connected only to the definition of kernel itself. First, we take different values for $(\sigma,t)$ and we plot the corresponding accuracies, here in the case of MUTAG with shortest path distance, but the same behavior holds true also for other datasets.

	The condition numbers are indeed high for every choice of parameters and therefore we avoid reporting here because it would be meaningless. From the plot, it is evident that it is convenient to set $\sigma$ lower or equal to 10 instead $t$ should be set bigger or equal to 0.1. Thus in what follows, we take into account $\sigma \in \{0.001,0.01,0.1,1,10\}$ and $t \in \{0.1,1,10, 100,1000\}$.
	
	In the case of \textbf{PI}, we considered a reasonable set of values for the parameter $\sigma \in \{0.001,0.01,0.1,1,10,100,1000\}$. The results are related to BZR with shortest path distance.
	
	\begin{figure}[h]
		\centering
		\includegraphics[scale=0.45]{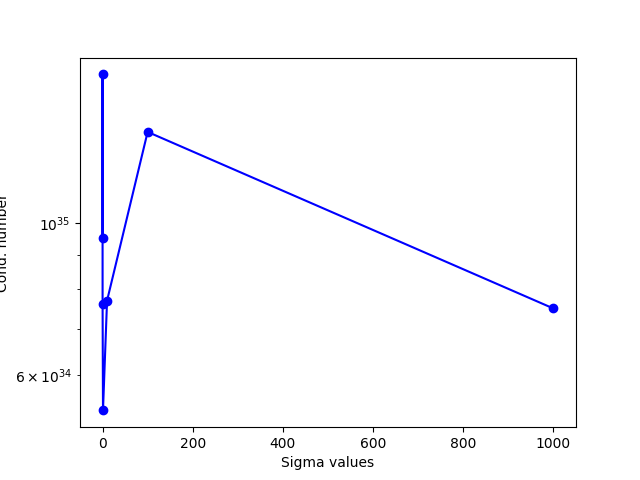} \quad \includegraphics[scale=0.45]{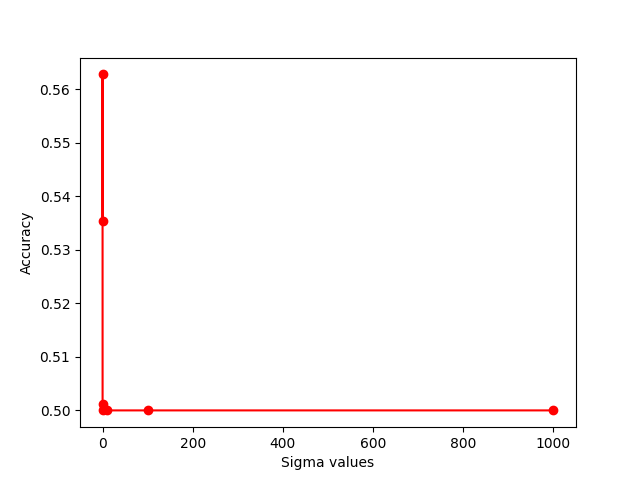}
		
		\caption{Comparison results about PI for BZR in terms of condition number (left) and accuracy (right) with different $\sigma$}
	\end{figure}	
	
	As in the previous kernels, it seems that the accuracy is better for small values of $\sigma$. For this reason, we set $\sigma \in \{0.000001,0.00001,0.0001,0.001,0.01,0.1,1,10\}$.
	
	\section{Numerical Tests}
	For what concerns the computation of simplicial complexes and persistence diagrams, we use some Python libraries available online as {\tt gudhi} \cite{gudhi}, {\tt ripser} \cite{ripser}, {\tt giotto-tda} \cite{giotto} and {\tt persim} \cite{persim}.
	To all datasets, we have performed a random splitting $(70\%/30\%)$ for training and test and applied a 10-fold Cross Validation on the training set for the hyperparameters tuning. Then we averaged the results over $10$ runs.
	For balanced datasets, we have measured the performances of classifier through the accuracy for binary and multiclass problems, 
	
	$$\text{accuracy} = \frac{\text{number of test samples correctly classify}}{\text{all test samples}}$$
	
	Instead, in the case of imbalanced datasets, we have adopted the balanced accuracy as explained in \cite{Metrics}, if for every class $i$ we define the related recall as
	
	$$recall_{i} = \frac{\text{test samples of class i correctly classify}}{\text{all test samples of class i}}$$
	
	then the balanced accuracy, in case of $n$ different classes is,
	
	$$\text{balanced\_accuracy} = \frac{\sum_{i=1}^{n} recall_{i}}{n}$$	
	
	This definition is able to effectively quantify how accurate is the classifier even in the case of the smallest classes.
	In tests, we use the implementation of SVM provided by the {\tt Scikit} \cite{SVMPython} library of Python. For PFK, we precomputed the Gram matrices using a Matlab (Matlab R2023b) routine because it is faster than the Python one. The values for $C$ belong to $\{0.001,0.01,0.1,1,10,100\}$. For each kernel, we have considered the following values for the parameters:
	\begin{itemize}
		\item \textbf{PSSK}: $\sigma \in \{0.00001,0.0001,0.001,0.01,0.1,1,10\}$
		\item \textbf{PWGK}: $\tau \in \{0.001,0.01,0.1,1,10,100\}$, $\rho \in \{0.001, 0.01, 0.1,1,10,100,1000\}$, $p \in \{1,5,10,50,100\}$, $C_{w} \in \{0.001,0.01,0.1,1\}$ and for kernel we chose the Gaussian one.
		\item \textbf{SWK}: $\eta \in \{0.00001,0.0001,0.001,0.01,0.1,1,10\}$
		
		\item \textbf{PFK}: $\sigma \in \{0.001,0.01,0.1,1,10\}$ and $t \in \{0.1,1,10, 100,1000\}$
		\item \textbf{PI}: $\sigma \in \{0.000001,0.00001,0.0001,0.001,0.01,0.1,1,10\}$ and number of pixel 0.1.
	\end{itemize}
	
	All codes have been run using Python 3.11 on a 2.5 GHz Dual-Core Intel Core i5, 32 Giga RAM. They are available in the GitHub page
	
	\begin{center}
		{\tt https://github.com/cinziabandiziol/persistence\_kernels}
		
	\end{center}
	
	\subsection{Point cloud data and shapes}
	\subsubsection{Protein}
	This is the Protein Classification Benchmark dataset PCB00019 \cite{Protein}. It sums up information for 1357 proteins corresponding to 55 classification problems. The data are highly imbalanced and therefore we apply the classifier to one of them, where the imbalance is slightly less evident.
	Persistence diagrams were computed for each protein by considering the 3-D structure or better the $(x,y,z)$ position of any atoms in each of the 1357 molecules, as a point cloud in $\mathbb{R}^{3}$. Finally using {\tt ripser} we compute the persistence diagrams only of dimension 1.
	
	\subsubsection{SHREC14 - Synthetic data}

    The dataset is related to the problem of non-rigid 3D shape retrieval. It collects exclusively human models in different body shapes and 20 poses. It consists of 15 different human models, about man, woman, and child, each with its own body shape. Each of these models exists in 20 different poses making the dataset composed of 300 models.
 
	\begin{figure}[h]
		\begin{center}
			\includegraphics[scale=0.75]{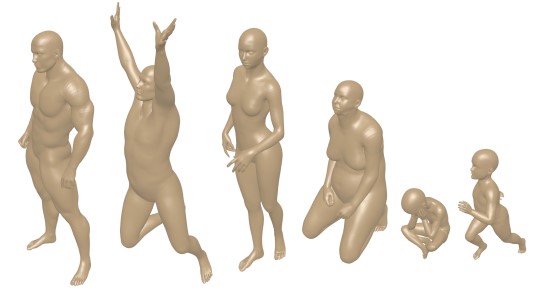}
		\end{center}
	\end{figure}
	 
	For each shape, the meshes are given with about 60000 vertices and, using the Heat Kernel Signature (HKS) introduced in \cite{HKS}, over different values of $t_{i}$ as \cite{PSSK}, we have computed the persistence diagrams of the induced filtrations in dimensions 1.
	
	\subsubsection{Orbit recognition}
	
	We consider the dataset proposed in \cite{PI}. We take into account the linked twisted map, which models fluid flows. The orbits can then be computed through the following discrete dynamical system
	\begin{equation*}
		\begin{cases}
			x_{n+1} &= x_{n} + ry_{n}(1-y_{n}) \text{ mod 1}\\
			y_{n+1} &= y_{n} + rx_{n+1}(1-x_{n+1}) \text{ mod 1}
		\end{cases}
	\end{equation*}
	where the starting point $(x_{0},y_{0}) \in [0,1] \times [0,1]$ and $r > 0$ is a real parameter that influences the behavior of the orbits, as appears in images.
	\begin{figure}[h]
		\includegraphics[scale=0.7]{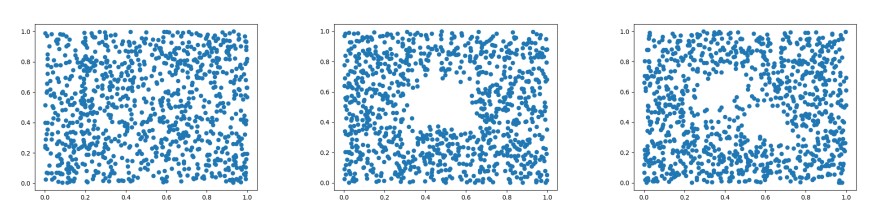}
		\caption{Orbits composed by the first 1000 iterations of the twisted map with $r = 3.5, 4.1, 4.3$ from left to right}
	\end{figure}
	
	As in \cite{PI}, $r \in {2.5,3.5,4,4.1,4.3}$ and it is strictly connected to the label of the corresponding orbit. For each of them, we compute the first 1000 points of 50 orbits, with starting points chosen randomly. The final dataset is composed of 250 elements. We compute PD considering only the 1-dimensional features. Since each PD has a huge number of topological features, we decide to consider only the first 10 most persistent, as done in \cite{VSPK}.
	
	\begin{table}[h]
		\begin{center}
			\begin{tabular}{|c|c|c|c|} \hline
				\textbf{Kernel} & \textbf{PROTEIN} & \textbf{SHREC14} & \textbf{DYN SYS}\\ \hline
				PSSK & \textbf{0.561} & 0.933 & 0.829\\
				\hline
				PWGK & 0.538 & 0.923 & 0.819\\ \hline
				SWK & 0.531 & \textbf{0.935} & \textbf{0.841}\\
				\hline
				PFK & 0.556 & \textbf{0.935} & 0.784\\ \hline
				PI & 0.560 & 0.934 & 0.777\\ \hline
			\end{tabular}
		\end{center}\caption{Accuracy related to point cloud and shape datasets}
	\end{table}
	
	First, the high difference in performances through different datasets is probably due to the high imbalance of the PROTEIN one with respect to the perfect balance of the other ones. It is well known that, if the classifier has not enough samples for each class, as in the case of the imbalanced dataset, it has to face high issues in classifying correctly elements of the minor classes. Except for PROTEIN where the PSSK shows slightly better performances, for SHREC14 and DYN SYS the best accuracy has been achieved by SWK.
	
	\subsection{Images}
	
	All the definitions introduced in Section 2 can be extended to another kind of simplicial complex, the cubical complex. It is useful when one deals with images or objects based on meshes, for example. More precise from \cite{Cubical}
	
	\begin{defn}
		An elementary cube $Q \subset \mathbb{R}^{d}$ is defined as a product $Q = I_{1} \times \dots \times I_{d}$ where each $I_{j}$ is either a singleton set $\{m\}$ or a unit-length interval $[m; m+1]$ for some integers $m \in \mathbb{Z}.$ The number k of the unit-length intervals in the product of Q is called the dimension of cube Q and we call Q a k-cube. If Q and $\bar{Q}$ are two cubes and $Q \subset \bar{Q}$, then Q is said to be a face of $\bar{Q}$. A cubical complex X in $\mathbb{R}^{d}$ is a collection of k-cubes $(0 \leq k \leq d)$ such that:
		\begin{itemize}
			\item every face of a cube in X is also in X;
			\item the intersection of any two cubes of X is either empty or a face of each of them.
		\end{itemize}
	\end{defn}
	
	\begin{figure}[h]
		\centering
		\includegraphics{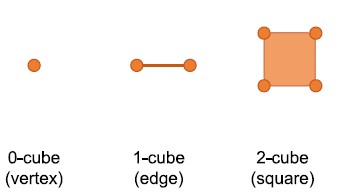}
		\caption{Cubical simplices}
	\end{figure}
	
	\subsubsection{MNIST and FMNIST}
	MNIST \cite{MNIST} is very common in the classification framework. It consists of 70000 handwritten digits, in grayscale, which one could try to classify into 10 different classes. Each image can be viewed as a set of pixels with a value between 0 and 256 (black and white) as in the figure. 
	
	\begin{figure}[h]
		\begin{center}
			\includegraphics[scale=0.95]{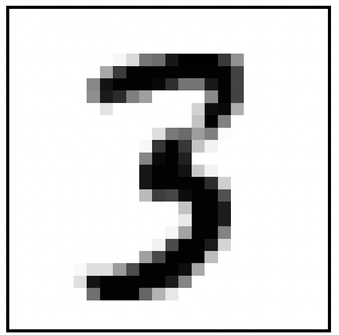}
		\end{center}\caption{Example of an element in MNIST dataset}
	\end{figure}
	
	Starting from this kind of dataset, we have to compute the corresponding persistent features. According to the approach proposed in \cite{TDA_MNIST} coming from \cite{Comp_study}, we first binarize each image, for instance, we replace each grayscale image with a white/black one, then we use as filtration function the so-called \textbf{Height filtration} $\mathcal{H}(p)$ in \cite{TDA_MNIST}. For cubical complex, for a chosen vector $v \in \mathbb{R}^{d}$ of unit norm, it is defined as
	
	\begin{equation*}\def\meno{\phantom{0}}
		\mathcal{H}(p) =
		\begin{cases}
			\langle p,v \rangle  & \text{if p is black},\\
			H_{\infty}  & \text{otherwise}
		\end{cases}
	\end{equation*}
	
	where $H_{\infty}$ is a big default value chosen by the user. As in \cite{Comp_study} we have chosen 4 different vectors for p: $(1,0),(-1,0),(0,1),(0,-1)$ and we have computed 0 and 1-dimensional persistent features both using {\tt tda-giotto} and {\tt gudhi} libraries. Finally, we concatenate them.
	For the current experiment, we decided to focus the test on a subset of the original MNIST, composed of only 10000 samples. This is a balanced dataset. Due to some memory issues, we have to consider for this dataset a pixel size of $0.5$ and for PWGK only $\tau \in \{0.001,0.01,0.1,1,10,100\}$, $\rho \in \{0.001, 0.1,10,1000\}$, $p = 10$, $C_{w} \in \{0.001,0.01,0.1,1\}$.
	
	Another example of a grayscale image dataset is the FMNIST \cite{FMNIST}, which contains 28 x 28 grayscale images related to the fashion world. 
	
	\begin{figure}[h]
		\begin{center}
			\includegraphics[scale=0.25]{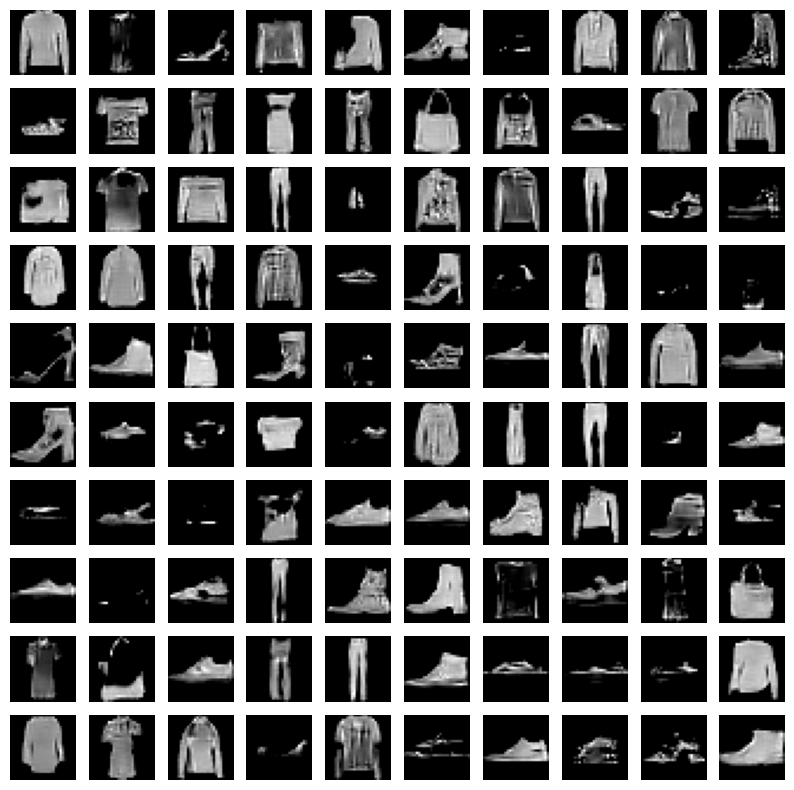}
		\end{center}\caption{Example of an element in FMNIST dataset}
	\end{figure}
	
	To deal with it, we follow another approach proposed in \cite{TDA_FMNIST}, where the authors apply padding, median filter, shallow thresholding, and canny edges and then compute the usual filtration to the image obtained. Due to some memory issues, we have to consider for this dataset a pixel size of $1$ and for PWGK only $\tau \in \{0.001,0.01,0.1,1,10,100\}$, $\rho \in \{0.001, 0.1,10,1000\}$, $p = 10$, $C_{w} \in \{0.001,0.01,0.1,1\}$.
	
	\begin{table}[h]
		\begin{center}
			\begin{tabular}{|c|c|c|} \hline
				\textbf{Kernel} & \textbf{MNIST} & \textbf{FMNIST}\\ \hline
				PSSK & 0.729 & 0.664\\ \hline
				SWK & \textbf{0.802} & \textbf{0.709} \\ \hline
				PWGK & 0.754 & 0.684\\ \hline
				PFK &  0.734 & 0.671\\ \hline
				PI &  0.760 & 0.651\\ \hline
			\end{tabular}\caption{Accuracy related to MNIST and FMNIST}
		\end{center}
	\end{table}
	
	Both datasets are balanced and probably the results are better in the case of MNIST due the fact that it is easier to classify handwritten digits instead of images of cloths. SWK shows slightly best performances.

	\subsection{Graphs}	
	
	In many different contexts from medicine to chemistry, data can have the structure of graphs. Graphs are couples of set $(V,E)$ where $V$ is the set of vertices and $E$ is the set of edges. The graph classification is the task of attaching a label/class to each whole graph. In order to compute the persistent features, we need to build a filtration. In the context of graphs, as in other cases, there are different definitions, see for example \cite{Graph-filt}. 
	
	We consider the Vietoris Rips filtration, where starting from the set of vertices, at each step we add the corresponding edge whose weights are less or equal to a current value $\epsilon$. This turns out to be the most common choice and the software available online allows us to build it after providing the corresponding adjacency matrix. In our experiments, we consider only undirected graphs but, as in \cite{Graph-filt}, building a filtration is possible also for directed graphs. Once defining the kind of filtration to use, one needs again to choose the corresponding weights. We decide to take into account: first, the shortest path distance and then the Jaccard index as for example in \cite{WKPI}.
	
	Given two vertices $u,v \in V$ the \textbf{shortest path distance} is defined as the minimum number of different edges that one has to meet going from u to v, or vice versa since the graphs here are considered as undirected. In graphs theory, it is a widely use metric.
	
	Instead, the Jaccard index is a good measure of edge similarity.
	Given an edge $e = (u,v) \in E$ then the corresponding \textbf{Jaccard index} is computed as
	
	$$\rho(u,v) = \bigg|\frac{NB(u) \cap NB(v)}{NB(u) \cup NB(v)}\bigg|$$
	
	where $NB(u)$ is the set of neighbours of $u$ in the graph. This metric recover local information of nodes in the sense that two nodes are considered similar if their neighbor sets are similar.
	
	In both cases, we consider the sub-level set filtration and we collect 0 and 1 dimensional persistent features both.
	
	We take 6 of such sets among the graph benchmark datasets, all undirected. They are
	\begin{itemize}
		\item \textbf{MUTAG}: it is a collection of nitroaromatic compounds and the goal is to predict their mutagenicity on Salmonella typhimurium
		\item \textbf{PTC}: is a collection of chemical compounds represented as graphs that report the carcinogenicity of rats
		\item \textbf{BZR}: it is a collection of chemical compounds and one has to classify them as active or inactive
		\item \textbf{ENZYMES}: it is a dataset of protein tertiary structures obtained from the BRENDA enzyme database and the aim is to classify each graph into 6 enzymes.
		\item \textbf{DHFR}: it is a collection of chemical compounds and one has to classify them as active or inactive
		\item \textbf{PROTEINS}: in each graph nodes represent the secondary structure elements and the task is to predict whether a protein is an enzyme or not.
	\end{itemize}
	
	Their properties are summarized in table \ref{Graph_label}, where the IR index is the so-called \textbf{Imbalanced Ratio} (\textbf{IR}), that denotes the imbalance of the dataset, and it is defined as, a sample size of the major class over sample size of the minor class.
	
	\begin{table}[h]
		\begin{center}
			\begin{tabular}{|c|c|c|c|} \hline
				\textbf{Dataset} & \textbf{N° Graphs} & \textbf{N° classes} & \textbf{IR}\\ \hline
				MUTAG & $188$ & $2$ & 125:63\\ \hline
				PTC & $344$ & $2$ & 192:152\\ \hline
				BZR & $405$ & $2$ & 319:86\\ \hline
				ENZYMES & $600$ & $6$ & 100:100\\ \hline				
				DHFR & $756$ & $2$ & 461:295\\ \hline
				PROTEINS & $1113$ & $2$ & 663:450\\ \hline
			\end{tabular}\caption{Graph datasets}\label{Graph_label}
		\end{center}
	\end{table}	
	
	Computations of adjacency matrix and PDs are made using functions implemented in {\tt tda-giotto}.
	
	The performances achieved with two edge weights are reported in tables,
	
	\begin{table}[h]
		\begin{center}
			\begin{tabular}{|c|c|c|c|c|c|c|} \hline
				\textbf{Kernel} & \textbf{MUTAG} & \textbf{PTC} & \textbf{BZR} & \textbf{DHFR} & \textbf{PROTEINS} & \textbf{ENZYMES}\\ \hline
				PSSK & 0.868 & \textbf{0.545} & 0.606 & 0.557 & 0.668 & 0.281  \\ \hline
				PWGK & 0.858 & 0.510 & 0.644 & 0.655 & \textbf{0.694} & 0.329 \\ \hline
				SWK & \textbf{0.872} & 0.511 &\textbf{0.712} & \textbf{0.656} & 0.686 & \textbf{0.370} \\ \hline
				PFK & 0.842  & 0.534 & 0.682 & \textbf{0.656} & \textbf{0.694} & 0.341 \\ \hline
				PI & 0.863 & 0.542 & 0.585 & 0.519 & 0.691 & 0.285 \\ \hline
			\end{tabular}\caption{Balanced Accuracy related to graph datasets using shortest path distance (Accuracy only for ENZYMES dataset)}
		\end{center}
	\end{table}
	
	\begin{table}[h]
		\begin{center}
			\begin{tabular}{|c|c|c|c|c|c|c|} \hline
				\textbf{Kernel} & \textbf{MUTAG} & \textbf{PTC} & \textbf{BZR} & \textbf{DHFR} & \textbf{PROTEINS} & \textbf{ENZYMES} \\ \hline
				PSSK & 0.865 & 0.490 & 0.704 & 0.717 & 0.675 & 0.298 \\ \hline
				PWGK & 0.859 & 0.516 & \textbf{0.720} & 0.727 & \textbf{0.699} & 0.355  \\ \hline
				SWK & 0.858 & 0.523 & 0.703 & 0.726 & 0.689 & \textbf{0.406} \\ \hline
				PFK & \textbf{0.874} & \textbf{0.554} & 0.704 & \textbf{0.743} & 0.678 & 0.400  \\ \hline
				PI & 0.846 & 0.478 & 0.670 & 0.712 & 0.690 & 0.280 \\ \hline
			\end{tabular}\caption{Balanced Accuracy related to graph datasets using Jaccard Index (Accuracy only for ENZYMES dataset)}
		\end{center}
	\end{table}
	
	Thanks to these results two conclusions can be taken. The first one is that, as expected, the performance of the classifier depends on the particular filtration used for the computation of persistent features. The second one is related to the fact that SWK and PFK seem to work slightly better than the other kernels: in the case of shortest path distance, SWK is to be preferred while PFK seems work better in the case of Jaccard index. In the case of PROTEINS, in both cases PWGK provides best balanced accuracy.
	
	\subsection{1-Dimensional Time Series}
	
	In many different applications, one can deal with 1-dimensional time series. A 1-dimensional time series is a set $\{x_{t} \in \mathbb{R} | t=1,\dots,T\}$. In \cite{Signals} authors provide different approaches to build a filtration upon this kind of data. We decide to adopt the most common one. Thanks to the Taken's embedding, these data can be translated into point clouds. With suitable choices for two parameters: $\tau > 0$ the delay parameter and $d > 0$ the dimension, it is possible to compute a subset of points in $\mathbb{R}^{d}$ composed by $v_{i} = \{x_{i}, x_{i+\tau}, \dots, x_{i+(d-1)\tau}\}$ for $i = 1, \dots, T - (d-1) \tau$. The theory mentioned above related to point clouds can now be applied to signals, as points in $\mathbb{R}^{d}$. For how to choose values for the parameters, see \cite{Signals}. The dataset for tests is taken from the UCR Time Series Classification Archive (2018) \cite{UCRSeries}, which consists of 128 datasets of time series from different worlds of application. In the archive there is the splitting into test and train sets but, for the aim of our analysis, we don't take care of it and we consider train and test data as a whole dataset and then codes provide properly the subdivision.
	
	\begin{table}[h]
		\begin{center}
			\begin{tabular}{|c|c|c|c|} \hline
				\textbf{Dataset} & \textbf{N° time series} & \textbf{N° classes} & \textbf{IR}\\ \hline
				ECG200 & $200$ & $2$ & 133:67\\ \hline
				SONY & $621$ & $2$ & 349:272\\ \hline
				DISTAL & $876$ & $2$ & 539:337\\ \hline
				STRAWBERRY & $983$ & $2$ & 632:351\\ \hline
				POWER & $1096$ & $2$ & 549:547\\ \hline				
				MOTE & $1272$ & $2$ & 685:587\\ \hline
				
			\end{tabular}\caption{Time series datasets}
		\end{center}
	\end{table}

	Using { \tt giotto-tda} we compute the persistent features of dimensions 0,1,2 and join them together. The final results of the datasets are reported here.
	
	\begin{table}[h]
		\begin{center}
			\begin{tabular}{|c|c|c|c|c|c|c|} \hline
				\textbf{Kernel} & \textbf{ECG200} & \textbf{SONY} & \textbf{DISTAL} & \textbf{STRAWBERRY} & \textbf{POWER} & \textbf{MOTE}\\ \hline
				PSSK & 0.642 & 0.874 & 0.658 & 0.814 & 0.720 & 0.618 \\ \hline
				PWGK & 0.726 & 0.888 & 0.696 & 0.892 & 0.769 & 0.633 \\ \hline
				SWK & \textbf{0.731} & 0.892 & \textbf{0.723} & \textbf{0.898} & \textbf{0.784} & \textbf{0.671} \\ \hline
				PFK & 0.707 & \textbf{0.895} & 0.676 & 0.892 & 0.750 & 0.652 \\ \hline
				PI & 0.717 & 0.841 & 0.662 & 0.793 & 0.712 & 0.606 \\ \hline
			\end{tabular}\caption{Balanced Accuracy related to time series datasets}
		\end{center}
	\end{table}
	
	As in the previous examples, SWK is winning and provides slightly best performances in terms of accuracy.
	
	\section{Conclusions}
In this paper, we have compared the performance of five Persistent Kernels applied to data of different natures. The results show how different PK are indeed comparable in terms of accuracy and there is not a PK that emerges clearly above the others. However, in many cases, the SWK and PFK perform slightly better. In addition, from a purely computational point of view, SWK is to be preferred since, by construction, the preGram matrix is parameter-independent. Therefore, in practice, the user has to compute such a matrix on the whole dataset only once at the beginning and then choose a suitable subset of rows and columns to perform the training, cross-validation, and test phases. This aspect is relevant and reduces the computational costs and time compared with other kernels.
Another aspect to be considered, as in the case of graphs, is how to choose the function $f$ that provides the filtration. The choice of such a function is still an open problem and an interesting field of research. The right choice in fact would guarantee to be able to better extract the intrinsic information from data, improving, in this way, the classifier's performances.
For the sake of completeness, we recall here that in the literature there is also an interesting direction of research whose aim is to build a new PK starting from the main 5. From one of the PKs mentioned in previous sections, the authors in \cite{VSPK} studied how to modify them obtaining the so-called \textbf{Variably Scaled Persistent Kernels}, which are 
Variably Scaled Kernels applied to the classification context. The results reported by the authors are indeed promising, thus it could be another interesting direction for further analysis.

	\vskip 0.1in
	
	{\bf Acknowledgments.} This research has been achieved as part of RITA \textquotedblleft Research
	ITalian network on Approximation'' and as part of the UMI topic group ``Teoria dell'Approssimazione
	e Applicazioni''. The authors are members of the INdAM-GNCS Research group. The project has also been funded by the European Union-Next Generation EU under the National Recovery and Resilience Plan (NRRP),  Mission 4 Component 2 Investment 1.1 - Call PRIN 2022 No. 104 of February 2, 2022 of Italian Ministry of University and Research; Project 2022FHCNY3 (subject area: PE - Physical Sciences and Engineering) \lq\lq Computational mEthods for Medical Imaging (CEMI)\rq\rq.

\end{document}